\documentclass[pmlr]{jmlr}

\usepackage{longtable}

\usepackage{booktabs}

\theorembodyfont{\upshape}
\theoremheaderfont{\scshape}
\theorempostheader{:}
\theoremsep{\newline}

\usepackage{graphics, color, graphicx, wrapfig}
\graphicspath{ {./fig/} }
\usepackage{float}
\usepackage{comment}
\usepackage{multirow}

\usepackage{makecell}

\usepackage{adjustbox}

\jmlrvolume{1}
\jmlryear{2022}
\jmlrworkshop{Workshop Title}

\title[Lessons learned from the NeurIPS 2021 MetaDL challenge]{Lessons learned from the NeurIPS 2021 MetaDL challenge: Backbone fine-tuning without episodic meta-learning dominates for few-shot learning image classification}

  \author{
    \Name{Adrian {El Baz}}\footnote{The two first authors are principal challenge organizer and dataset preparer; the other authors are in alphabetical order.} \Email{eb.adrian8@gmail.com}\\
    \Name{Ihsan Ullah} \Email{ihsan2131@gmail.com}\\
    \Name{Edesio Alcoba\c{c}a} \Email{e.alcobaca@gmail.com}\\
    \Name{Andr\'{e} C. P. L. F. Carvalho} \Email{andre@icmc.usp.br}\\ 
    \Name{Hong Chen} \Email{h-chen20@mails.tsinghua.edu.cn} \\
    \Name{Fabio Ferreira} \Email{ferreira@cs.uni-freiburg.de}\\
    \Name{Henry Gouk} \Email{henry.gouk@ed.ac.uk}\\
    \Name{Chaoyu Guan} \Email{guancy19@mails.tsinghua.edu.cn}\\
    \Name{Isabelle Guyon} \Email{guyon@chalearn.org}\\
    \Name{Timothy Hospedales} \Email{t.hospedales@ed.ac.uk}\\
    \Name{Shell Hu} \Email{shell.hu@samsung.com}\\
    \Name{Mike Huisman} \Email{m.huisman@liacs.leidenuniv.nl}\\
    \Name{Frank Hutter} \Email{fh@cs.uni-freiburg.de}\\
    \Name{Zhengying Liu} \Email{zhengying.liu@inria.fr}\\
    \Name{Felix Mohr} \Email{felix.mohr@unisabana.edu.co}\\
    \Name{Ekrem \"Ozt\"urk} \Email{ozturk@informatik.uni-freiburg.de}\\
    \Name{Jan N. van Rijn} \Email{j.n.van.rijn@liacs.leidenuniv.nl}\\
    \Name{Haozhe Sun} \Email{haozhe.sun@universite-paris-saclay.fr}\\
    \Name{Xin Wang} \Email{xin\_wang@tsinghua.edu.cn}\\
    \Name{Wenwu Zhu} \Email{wwzhu@tsinghua.edu.cn}\\
  }

\editor{Editor's name}

\begin{document}

\maketitle

\begin{abstract}
Although deep neural networks are capable of achieving performance superior to humans on various tasks, 
they are notorious for requiring large amounts of data and computing resources, restricting their success to domains where such resources are available.
Meta-learning methods can address this problem by transferring knowledge from related tasks, thus reducing the amount of data and computing resources needed to learn new tasks. 
We organize the MetaDL competition series, which provide opportunities for research groups all over the world to create and experimentally assess new meta-(deep)learning solutions for real problems.
In this paper, authored collaboratively between the competition organizers and the top-ranked participants, we describe the design of the competition, the datasets, the best experimental results, as well as the top-ranked methods in the {\bf NeurIPS 2021 challenge}, which attracted 15 active teams who made it to the final phase (by outperforming the baseline), making over 100 code submissions during the feedback phase. The solutions of the top participants have been {\bf open-sourced}. 
The lessons learned include that learning good representations is essential for effective transfer learning.

\end{abstract}

\section{Introduction}

Automated Machine Learning (AutoML) has made great progress in the past years, driven by the increasing need for solving machine learning and data science tasks efficiently with limited human effort.
A variety of approaches have been proposed, including Autosklearn (using Bayesian optimization, \citet{feurer2015robust}), MLPlan (using tree Search, \citet{mohr2018mlplan}), TPOT (using genetic algorithms, \citet{olsonM19tpot}), and learning curves~\citep{mohr2022learning}.

Machine learning challenges have been instrumental in benchmarking AutoML methods~\citep{guyon2019analysis}, and bringing to the community state-of-the-art, open-source solutions, such as auto-sklearn~\citep{feurer2015robust}. Recently, AutoML challenges have started addressing deep learning problems \citep{liu2020towards,liu_2021} and it has become clear that meta-learning is one central aspect that deserves more attention. Meta-learning~\citep{brazdil2022metalearning} aims at `learning to learn' more effectively and transferring expertise from task to task, to improve performance, cut down training times and the need for human expertise, and/or reduce the number of training examples needed. 

Advances in meta-learning are beneficial for a wide range of scientific domains, in particular when obtaining a large number of examples of any given task is costly. 
Such problems are often encountered, e.g., in medical image analysis, diagnosis of rare diseases, design of new materials and analysis of questionnaires.
Other areas in which meta-learning is particularly critical include multi-class classification problems in which some classes are particularly rare, to the extent that only one or two examples may be available, e.g., rare plant or animal species. 
Such applications of meta-learning, focusing on rare, expensive, or difficult data to collect, have obvious economic and societal impact. Furthermore, work on meta-learning will contribute to advancing methods, which reduce the need for human expertise in machine learning
and democratize the use of machine learning (by open-sourcing automated methods). 
In this challenge, we have aimed to maximize the societal impact of our effort by assembling datasets from a variety of practical domains, with direct relevance to `AI for good', including medicine, ecology, biology, and pharmacology. 

Our contributions are the following. 
We have developed a new benchmark for meta-learning, consisting of 10 new datasets from 5 practical domains, which we aim to make publicly available. 
The challenge has been entered by 15 teams,\footnote{See: \url{https://autodl.lri.fr/competitions/210}, this excludes test submissions made by various organizers} who made a total of more than 100 submissions.
The top-ranked method, MetaDelta++~\citep{metadelta}, was able to achieve an accuracy score of more than $0.92$ on all of the 5 hidden datasets.

\section{Competition Setup}
In this section, we describe the framework, the datasets and the implementation of the challenge.

\subsection{Framework}
The competition follows the protocol of~\citet{elbaz2021metadl}, aiming at testing models on {\bf few-shot learning problems}, in the `N-way k-shot setting'. The main differences in this new competition, are that more and harder datasets are provided, the number of shots (examples per class) is increased from 1 to 5, while the number of ways (classes per task) remains constant at 5. 

The competition spans two phases: (1) a {\bf feedback phase} (in which participants iteratively develop and submit their methods and get immediate performance feedback on 5 meta-datasets hidden on the platform by submitting their code), and (2) a {\bf final phase} (in which the last method of each participant submitted in the feedback phase is evaluated on 5 fresh final meta-datasets). 
The feedback phase lasted for 2 months, after which the final phase was run to determine the final ranking of the participants, using the final submission of the participants, in a fully blind test (no feedback). 

This competition focuses on few-shot image classification tasks. We use the $N$-way, $k$-shot classification setting, commonly used in the meta-learning literature~\citep{finn2017model,snell2017prototypical,huisman2021survey}. The setup includes two stages: meta-training and meta-test. Given a (meta-)dataset, two disjoint class pools are created, one class pool is used for meta-training while the other one is used for meta-test. 
Each stage contains (possibly overlapping) tasks/episodes, each of which contains exactly $N$ classes (called ``ways'') and $k$ examples per class (called ``shots''). 
A meta-training set includes examples of such tasks (pairs of training and test sets; both labelled). 
A meta-test set is then presented to the meta-trained algorithms, including labelled training sets and unlabeled test sets. Since the meta-test classes were not seen during training, this is a form of out-of-distribution evaluation~\citep{setlur2021two}.

Each submitted script follows a specific API (as outlined by~\citet{elbaz2021metadl}), which we defined as challenge organizers. 
This API is designed to be flexible enough to be used to describe any meta-learning procedure. 
It defines 3 main classes which methods need to override to completely define a meta-learning algorithm. 
Its design relies on the definition of the different algorithms' implementation levels that have been identified by \citet{liu2019overview}. 
The 3 main classes are the following (as described by~\citet{elbaz2021metadl}):

\begin{itemize}
    \item \textbf{Meta-learner}: has a \texttt{meta\_fit()} method that encapsulates the meta-training procedure. Using the previously defined notation, it essentially processes the meta-dataset and captures the reusable information across meta-training tasks. It takes the meta-train set as an argument and outputs a \textbf{learner}.
    \item \textbf{Learner}: has a \texttt{fit()} method that encapsulates the training procedure (e.g., the adaptation phase). It takes a train set (examples and labels) as an argument  along with the associated information 
    from the meta-learning procedure to output a \textbf{predictor}.
    \item \textbf{Predictor}: has a \texttt{predict()} method that predicts the labels of test examples. It takes a test set (unlabeled examples) as an argument and returns the predictions. These can be evaluated by the competition software.
\end{itemize}

\subsection{Datasets}

We collected 10 datasets from 5 domains (two per domain). 
The meta-dataset of each phase consists of 5 datasets, one from each domain.
This ensures that there is some resemblance between the datasets that were public during the feedback phase and the datasets that the systems were evaluated on. 

The datasets used in this competition come from five domains: ecology, bio-medicine, manufacturing, optical character recognition (OCR) and remote sensing, each having two datasets. The datasets are image datasets with at least 20 classes and 40 images per class. Sample images from each datasets are shown in Figure~1 (in Appendix~A in the supplement). 

All datasets, except OCR, are preprocessed, i.e., cropped, resized with an anti-aliasing filter into 128x128 size, and in some cases, some background padding was added to make the images fit in a square. The OCR datasets are generated directly in the required size by the OmniPrint software~\citep{sun2021omniprint}. Details about the domain, number of classes/categories and number of images in each dataset along with the phase of the competition in which the datasets have been used is shown in Table~1 (in Appendix~A in the supplement). These datasets (and others in preparation for our next challenge) will be released as part of a benchmark suite called Meta-Album, which will be made publicly available. 

\subsection{Implementation}

The challenge is implemented on the CodaLab platform, which allows competitions with code submission and the implementation of flexible protocols. All submissions from the feedback phase were automatically run on CodaLab so that all participants were allocated an equal amount of computing resources on the same hardware. Once the feedback phase was completed, we did manually run the last valid submission of each participant on the meta-datasets associated with the final phase, on a dedicated CodaLab instance.

Each submission would run for a maximum of 10 hours (2 hours maximum per meta-dataset). The number of submissions was loosely constrained during the feedback phase (5 per day, 100 in total). However, in the final test phase, a single submission per team was allowed, thus preventing teams from obtaining feedback on their performance on the private test datasets. 

In the feedback phase, scores on the 5 feedback phase datasets were publicly visible on the leaderboard, but the feedback phase datasets themselves (and their identity) were not accessible to the participants. The final evaluation was carried out in similar conditions on the final phase datasets. The final datasets and their identity also remained inaccessible to the participants.

NeurIPS 2021 MetaDL competition was built upon the infrastructure of the previous AAAI 2021 MetaDL competition. We provided participants with the same starting package that we used for the AAAI 2021 competition.\footnote{\url{https://github.com/ebadrian/metadl/tree/master/starting_kit}} It contains baseline methods such as MAML~\citep{finn2017model}, Prototypical networks~\citep{snell2017prototypical} and a naive baseline which accumulates all data from meta-training and trains a neural network on it. 

\section{Top ranked participants}
All participants were invited to fill out fact sheets. Top-ranked teams summarized theirs below and were invited to co-author this paper.

\subsection{MetaDelta++}
The authors of MetaDelta++ (displayed as team `ForeverYong' in the ranking of Table~\ref{tab:results}) won the challenge. They enhanced their previous MetaDelta solution, which won the AAAI 2021 MetaDL-mini competition~\citep{metadelta}. It is composed of three base meta-learners with a pre-trained backbone at multi-scale input and different training strategies. 
Specifically, for each domain (since all domains are meta-trained and meta-tested separately) the parameters of the pre-trained backbone are adapted for each meta-learner to the specific meta-training subset (including samples from a subset of the classes in the given meta-dataset, the other classes being used for meta-testing). To that end, the authors remove the original last layer and substitute it with a 3-layer MLP classifier, then fine-tune the backbone by freezing the parameters of the shallower layers (layers close to the input) to preserve the general knowledge and prevent overfitting to the small meta-training dataset. Furthermore, hand-crafted data augmentation (like rotation) is designed to help the fine-tuning process. After obtaining a fine-tuned backbone, for each meta-learner, the three-layer MLP is removed and replaced by a prototype-based classification method during meta-testing (inference stage), modelled as an optimal transport problem~\citep{opt1}, based on the features extracted by the backbone. Finally, the three meta-learners are `auto-ensembled' to stabilize the performance of the whole meta-learning system, by which we mean that the best backbone is selected by (meta-)cross-validation. They carefully monitor the time budget with a well-designed controller.

Ablation studies, conducted by the authors, show that the pre-trained backbones and the specially designed fine-tuning methods are of great significance for few-shot learning, while data augmentation and optimal transport can also help but not so critically.

The authors made the solution available on GitHub.\footnote{\url{https://github.com/Frozenmad/MetaDelta}}

\subsection{Edinburgh-Samsung: Self-supervised transfer learning}
Following recent debate about simple embedding-learning {\em vs.} meta-learning \citep{chen2019closer,tian2020rethinking,zhang2021metaQDA}, this team (displayed as team `henrygouk' in the ranking of Table~\ref{tab:results}) eschewed meta-learning altogether and focused exclusively on leveraging larger external pre-training datasets and neural architectures. In particular, due to a previous observation showing self-supervised features often obtain better performance in transfer learning settings \citep{ericsson2021well}, they ultimately exploited the powerful vision transformer architecture trained with self-supervision by DINO \citep{caron2021emergingDINO} on ImageNet1K. The use of this substantial external dataset allowed fitting the larger ViT model. Their entry simply extracted DINO/ViT features and trained a well regularised logistic regression classifier during the meta-test step. 

The simplicity and comparatively high performance of this solution demonstrate that large scale pre-training and modern architectures may often be the easiest way to achieve high-performance few-shot learning in practice. A refined version of this approach is under review \citep{hu2020pmt}. The code for this entry is available on GitHub.\footnote{\url{https://github.com/henrygouk/neurips-metadl-2021}}

\subsection{Meta-Padawan}
The Meta-Padawan (displayed as team `padawan' in the ranking of Table~\ref{tab:results}) 
solution learns from previous pre-trained models, combining features descriptors previously learned to achieve fast model generalization using few images. The hypothesis was that it is possible to rapidly generalize new deep neural network models by combining different features descriptors learned from previous models with simple features extracted from images themselves. To assess this hypothesis, Meta-Padawan designed the 2SoF  (Two Set of Features) method, which combines features from pre-trained models and images. Initially, 2SoF does data augmentation using training images, generating 10 new images for each image. This uses a rotation range of 20, zoom range of 0.15, width/height shift of 0.2, the sheer intensity of 0.15, and horizontal flip. 

Afterwards, 2SoF extracts features from each augmented image using Principal Component Analysis (PCA), capturing 95\% of the variance. These features complement the features extracted from the last convolutional layer from pre-trained models InceptionResNetV2 and VGG12~\citep{simonyan2014very, szegedy2017inception}. The combination of the PCA-based features with the features extracted by the last convolutional layers of deep networks creates a new feature space. Finally, the proposed method fits a Logistic Regression classification model with L2 regularization to this new feature space. No systematic ablation studies were conducted, but the authors confirm that both types of features contributed to the good performance of their method.

The code of Meta-Padawan was made publicly available on GitHub.\footnote{\url{https://github.com/ealcobaca/meta-padawan}}

\subsection{Meta-ZAP}
The authors of Meta-ZAP (Meta Zero-shot AutoML from Pre-trained Models, displayed as team `ekremtzr' in the ranking of Table~\ref{tab:results}) adapted their method from the AutoDL competition~\citep{liu_2021}, in which they meta-learn a model that selects the proper meta-learning pipeline, i.e., a pre-trained feature extractor and the hyperparameters of the meta-learning pipeline, based on the properties of a given dataset (e.g., the number of images and classes) in a zero-shot setting (without any model validation).

For learning this model selector, they create a meta-dataset that consists of classification accuracies (measured as in the competition setting) of many meta-learning pipelines across datasets and learn it with the algorithm selection approach AutoFolio \citep{lindauer_autofolio_2015}.

Contrary to their previous work in the AutoDL competition \citep{liu_2021}, they adapt their approach to the MetaDL setting by using a different configuration space consisting of seven hyperparameters to search for a meta-model and by fixing hyperparameters on fine-tuning~\citep{Saikia2020_few_shot}. Moreover, they generate 21 datasets via ICGen \citep{stoll2020_icgen}, a dataset-level augmentation tool, from several TensorFlow Datasets~\citep{tensorflow2015-whitepaper, TFDS} (TFDS) datasets and also include eight datasets from meta-dataset~\citep{triantafillou2020meta-dataset}. They utilize Task2Vec~\citep{AchilleLTRMFSP19} embeddings as dataset features. Additionally, they populate the source and support examples by applying image augmentations via TrivialAugment~\citep{Muller_2021_ICCV}. 

The Meta-ZAP solution was open-sourced on GitHub.\footnote{\url{https://github.com/ekremozturk/ZAP-few-shot}}

\section{Results}
In this section, we overview the results of the final phase and describe the method of how the individual scores are aggregated to a final ranking. 
\subsection{Final phase overview}
During the final phase, the last valid feedback phase submission from participants was considered and re-run from scratch on 5 new meta-datasets. As in the feedback phase, every submission is ranked on each meta-dataset independently and the final score of the submission is the average of these ranks. A submission from the feedback phase is considered valid if it beats the determined baseline using a popular few-shot learning method: Prototypical Networks \citep{snell2017prototypical}. 
A few participants had multiple submissions that beat the baseline, but only their last feedback phase submission was considered. Ultimately we had 15 participants entering the final phase.

\subsection{Evaluation protocol and results}
The 5 new meta-datasets in the final phase had similar image domains to the ones used during the feedback phase. To increase reproducibility, and avoid participants winning by chance, we ran each submission 3 times, using different random seeds. For each meta-dataset, we considered only the lowest accuracy among these 3 runs. The participants' results are displayed in Table~\ref{tab:results}.

\begin{table}[t]
    \centering
    \resizebox{\textwidth}{!}{
        \begin{tabular}{lrrrrrrrrrrr}
            \toprule
            & & \multicolumn{2}{c}{\textbf{Meta-dataset 1}} &
            \multicolumn{2}{c}{\textbf{Meta-dataset 2}} & 
            \multicolumn{2}{c}{\textbf{Meta-dataset 3}} & 
            \multicolumn{2}{c}{\textbf{Meta-dataset 4}} &
            \multicolumn{2}{c}{\textbf{Meta-dataset 5}}
            \\
            
             \textbf{Team}& \textbf{Average rank}& Accuracy& Rank& Accuracy& Rank& Accuracy& Rank& Accuracy& Rank& Accuracy& Rank \\
             \hline
             \textbf{ForeverYong} & 3.80 & 0.983 &2& 0.943 &\textbf{1}& 0.990 &\textbf{1}& 0.921 &\textbf{1}& - (0.939*) &14 (1*) \\
             \textbf{henrygouk} & 4.00& 0.979 &4& 0.770 &2& 0.419 &9& 0.803 &4& 0.875 &\textbf{1} \\
             \textbf{padawan} & 4.00& 0.981 &3& 0.714 &4& 0.488 &7& 0.856 &3& 0.865 &3 \\
             pikachu & 4.20 & 0.985 &\textbf{1}& 0.611 &6& 0.355 &10& 0.904 &2& 0.873 &2 \\
             ekremtzr & 4.60 & 0.937 &8& 0.732 &3& 0.921 &2& 0.762 &6& 0.817 &4 \\
             sheling343 & 5.20 & 0.977 &5& 0.659 &5& 0.515 &6& 0.778 &5& 0.783 &5 \\
             paisiasach & 6.20 & 0.949 &6& 0.412 &8& 0.684 &4& 0.698 &7& 0.732 &6 \\
             BucketHead & 7.20 & 0.946 &7& 0.411 &9& 0.786 &3& 0.660 &8& 0.639 &9 \\
             lucia & 7.80 & 0.842 &10& 0.467 &7& 0.552 &5& 0.612 &9& 0.698 &8 \\
             ericlhan & 8.80 & 0.851 &9& 0.382 &10& 0.458 &8& 0.611 &10& 0.719 &7 \\
             perathem & 11.20 & 0.749 &12& 0.334 &11& 0.208 &12& 0.411 &11& 0.538 &10 \\
             brunosez & 12.00 & 0.752 &11& 0.330 &12& 0.206 &14& 0.409 &12& 0.516 &11 \\
             vermashreth & 13.20 & 0.453 &14& 0.285 &13& 0.235 &11& 0.307 &14& 0.304 &14 \\
             Vilupa & 13.60 & 0.422 &15& 0.247 &14& 0.207 &13& 0.316 &13& 0.339 &13 \\
             mrm & 14.00 & 0.563 &13& 0.244 &15& 0.20 &15& 0.261 &15& 0.414 &12 \\
            \bottomrule
        \end{tabular}
    }
    \caption{Final phase results of the NeurIPS 2021 MetaDL competition. For each meta-dataset, the result displayed is the worst out of 3 runs. Again, for each meta-dataset, each algorithm is evaluated on 600 episodes with a 5-way 5-shot configuration. Classes and associated images are drawn from the associated meta-test set. The Accuracy columns are the average accuracy over the 600 episodes of the corresponding meta-dataset. *ForeverYong 5th meta-dataset result is displayed for completeness when the minor bug encountered is fixed.}
    \label{tab:results}
    \vspace{-0.2cm}
\end{table}

We evaluated several metrics to rank scores of each submission across meta-datasets: average rank, relative difference and the Copland method. Regardless of the metric used we obtained the same top-4 results. The choice between these metrics is tightly linked to the goal of the problem which usually is either to have a `generalist' or `specialist' algorithm \citep{pavao2021aircraft}.  The competition aims to foster algorithms that are capable of quickly dealing with few-shot image classification problems within a single image domain. Hence, we preferred the average rank metric which emphasized generalist behaviour. 

MetaDelta++ (team indicated as `ForeverYong') finished in first place with an improvement of the algorithm that got them the first place in the first edition of MetaDL at AAAI 2021 \citep{elbaz2021metadl}. Most notably, they did reach first place even though their algorithm encountered an error during the evaluation of the 5th meta-dataset, which resulted in the worst rank for the associated meta-dataset rank. The minor error was mainly due to a hardcoded number of classes to draw for episode generation, which worked well for the feedback phase but did not for the final phase. The next version of MetaDL will explicitly disclose final phase metadata to the participants to avoid these unfortunate errors. Also, there was a tie between the second and the third place and our competition rules state that in the occurrence of such an event, the algorithm that was submitted first is considered better. Therefore, Edinburgh-Samsung (displayed as `henrygouk') finished in the second position, while Meta-Padawan (displayed as `padawan') finished in the third position.

\section{Discussion}

In designing a good meta-learning system, several decision must be made:
\begin{enumerate}
    \item whether to use a {\bf filter} method to perform model selection and hyper-parameter selection (referred to as zero-shot learning by team Meta-ZAP) or a {
    \bf wrapper} method (with meta-cross-validation) like MetaDelta++, and whether {\bf meta-features} (extracted or provided) should be used to help model selection;
    \item whether to use a {\bf pre-trained backbone}, and if so, whether to {\bf fine-tune} its weights and how (with or without the {\bf provided meta-datasets} and with or without {\bf data augmentation};
    \item whether to use other types of {\bf feature extraction/embeddings} instead or in conjunction with features extracted from a pre-trained backbone;
    \item which type of {\bf classifier} to use after feature extraction, amenable to few-shot learning (e.g., simple linear classifier or example-based methods);
    \item whether and how to use model {\bf ensembles};
    \item how to efficiently manage the {\bf time budget}.
\end{enumerate}

Regarding filter {\em vs.} wrapper, in this challenge, the wrapper setting used by MetaDelta++ seems to have been very effective. It is more computationally expensive than filter methods, but the winning team made efficient use of the computational resources and therefore could afford it. Possibly a smart combination of filter and wrapper methods would be advisable, e.g., using filter methods as initialization or prior in a smart search strategy inspired by Bayesian optimization. This was not explored by the participants. Some participants attempted to compute dataset meta-features and noted that the organizers did not provide such meta-features, such as application domain, the scale of the image, etc. In particular, Meta-Padawan suggested providing a wider variety of image subtopics and pre-trained models to enhance the initial meta-knowledge. 

Regarding the use of pre-trained backbones, a complete consensus emerged: they are essential. 
All top-ranking participants used convolutional backbones or transformer models.
The winners MetaDelta++ managed to efficiently fine-tune the backbones they chose, but they warned about the danger of overfitting the meta-training data. 
To avoid that, they froze the layers closest to the input and used data augmentation. 
Interestingly, the second-place participants did not fine-tune their backbone. 
It was trained on ImageNet with self-supervised learning. 
Massive pre-training with large unlabeled datasets may be the solution of the future for image classification. 
However, this strategy relies on the fact that massive out-of-domain data is available. 
The value of small amounts of in-domain data vs large amounts of out-of-domain data (or how to combine both) remains to be seen.
In future challenges, we intend to have a dedicated track for comparing with de-novo training methods, which can only rely on the data provided. 

Another consensus is that `episodic' training (like MAML) is not necessary. 
Backbones are fine-tuned with regular gradient descent and classifiers are trained with standard classification algorithms. 
Several teams used a simple linear classifier, trained e.g., with logistic regression. 
The winners MetaDelta++ favoured a prototype-based method using optimal transport to compute a discrepancy metric rather than Euclidean distance. 
This finding is corroborated by other studies which demonstrate the benefit of access to global class labels~\citep{wang2021bridging,wang2021role} and show that classical episodic training can lead to suboptimal performance~\citep{laenen2021on}.
 
Finally, the optimization of usage of computational resources played an important role in winning, as reported by the participants. However, it is difficult to assess since it is entangled with algorithm implementation.  In particular, the competitors of the Meta-Padawan team shared that efficient time management and code were critical for them. They reported that processing costs could be significantly reduced by limiting data augmentation or avoiding unnecessary code loops.

From an organizational perspective, there were several pitfalls in our competition protocol that we will correct in the future.
The first thing that stands out from the results is that the MetaDelta++ solution (Team `ForeverYong' in Table~\ref{tab:results}) performs best on all public feedback datasets and on four out of five private test datasets, but {\em did not successfully run on the fifth dataset}. 
Analyzing the code, we discovered that this was due to a special case that was not anticipated.
In the future, we will provide more meta-data to facilitate anticipating such cases or a variety of datasets in the feedback phase covering all scenarios encountered in the final phase.
There is also room for improvement in our scoring method. Although the ranking of top participants was stable concerning the various possible methods of combining the 5 scores (each corresponding to a domain), it is known that the average rank method we used is influenced by the removal or addition of participants.
When a team submits a solution that crashes on a single dataset, as a consequence it will be ranked last. 
The severity of this penalty depends now on the number of participants; when there are only few participants this might not be that severe, whereas when there are many participants, this can be quite severe.
One possible remedy could be to introduce a maximum rank that participants can achieve so that one dataset does not extremely influence the score. 
We might also conduct bootstrap experiments to study the stability of ranking and report distributions rather than single results (see, e.g., \citet{pmlr-v133-turner21a}).

\section{Conclusion and further work}

We organized MetaDL (Meta-Deep-Learning), a competition to benchmark state-of-the-art meta-learning techniques and to further improve the field, focusing on few-shot image classification. 
The MetaDelta++ system~\citep{metadelta}, based on pre-trained backbone networks, performed best on most of these datasets. 
Other well-performing solutions are Edinburgh-Samsung~\citep{hu2020pmt}, Meta-Padawan and Meta-ZAP, which are all described in this paper. 
All winning solutions are open-sourced.

This competition is part of a well-established competition series, consisting of (among others) the AutoML and AutoDL competition series. 
Although it does not specifically constrain participants to use deep-learning, {\em de facto}, all participants based their solutions on deep-learning models with convolutions (specifically, either convolutional neural networks or transformer models).
Fine-tuning on meta-training data turned out to be important, though there are indications that off-the-shelf backbones pre-trained with self-supervised learning on massive datasets might become the way of the future, essentially making meta-learning unnecessary for image classification problems. 
Thus, meta-learning should be benchmarked in {\em de novo} training conditions, in the future, to prepare for scenarios (in other domains) in which such backbones are not available.

In this competition, we have refined the competition protocol developed for a previous MetaDL competition~\citep{elbaz2021metadl}, introducing multiple domains, which added additional sophistication in terms of scoring as well as GPU-time budgeting. 
However, our setting remained relatively simplified, in that images were small (128x128), meta-training and testing were performed within sub-domains (e.g., insect classification, texture classification, OCR, etc.), with sub-tasks involving only 5 classes (ways) relatively well separated (not from the same super-category) and with 5-shots. 
Indeed, the winners obtained over 92\% accuracy on all 5 domains in the final phase (with complete blind-testing of their code). 
Thus, we must move on to harder problems. In an upcoming challenge, we intend to mix tasks from multiple domains and present variable numbers of `ways' and `shots'. 
The participants also expressed the desire that we would organize an even more challenging competition, in the spirit of AutoDL \citep{liu_2021}, with meta-datasets stemming not only from different domains but different modalities (speech, image, video, text, tabular data, etc.) We might do this next.

For this competition, we have assembled 10 new datasets, from 5 practical domains.
From each domain, one dataset was used in a public feedback phase, and the other dataset was used in a private test phase. 
As reported by several top-ranking teams, meta-learning was possible within sub-domains (in the form of fine-tuning pre-trained backbone networks), but MAML-style episodic meta-learning did not turn out to be more effective than vanilla pre-training with gradient descent. 
Based on the embedding generated by the backbones, prototypical classifiers seem more efficient than linear classifiers. 
To further probe the effectiveness of various meta-learning solutions, we are preparing a larger cross-domain meta-learning challenge. 
We intend to re-design our API to facilitate ablation studies by modularizing the meta-learning workflow.

\subsubsection*{Acknowledgements} This paper is authored collaboratively by both the organizers and top-ranked participants of the competition. 
All participants that ranked in the top-5, and also filled in the fact-sheet detailing their method, were invited and co-authored the paper. 
Funding and support have been received by ANR Chair of Artificial Intelligence HUMANIA ANR-19-CHIA-0022, TAILOR (a project funded by EU Horizon 2020 research and innovation program under GA No. 952215) and ChaLearn, computing cloud units were donated by Microsoft and Google, and prizes were donated by 4Paradigm. 
We want to thank everyone that contributed to the creation of datasets: Jennifer (Yuxuan) He, Benjia Zhou, Professor Yui Man Lui, and Phan Anh Vu. We also thank  Sébastien Treguer and Adrien Pavao for helpful discussions.

\nocite{*}
\bibliography{competition-report}
\clearpage
{

\maketitle

\appendix

\section{Datasets}
\label{app:datasets}
Table~\ref{tab:datasets_summary} summarizes the datasets. Figure~\ref{fig:sampleimages} shows sample images per dataset.

\begin{table}[ht!]
  \caption{MetaDL 2021 datasets summary}
  \label{tab:datasets_summary}
  \centering
  \begin{adjustbox}{width=\linewidth}
  \begin{tabular}{l l l r r l r}
  
    \toprule
    \textbf{Domain}  & \textbf{Dataset} & \textbf{\makecell[l]{Competition\\Phase}}    & \textbf{Categories}   & \textbf{Images}  & \textbf{Source} 
    \\
    \midrule
    
    \multirow{2}{*}{Ecology}  
    &   Plankton & Feedback & 91  & 3,640   & \citet{whoiplankton}
    \\
    \cmidrule{2-6}
    &   Insects & Final & 114  & 4,560 & \citet{insects}
    \\
    \midrule
    
    \multirow{3}{*}{Bio-medicine} 
    &Multiderma & Feedback    &51 &2,040 &\citet{sdpaper}
    \\
    \cmidrule{2-6}
    &Plant Village & Final &37 &1,480  & \makecell[l]{\citet{hughes2012open}\\ \citet{geetharamani2019identification}}
    \\
    \midrule
    
    \multirow{5}{*}{Manufacturing} 
    &   Texture DTD & Feedback &  47  & 1,880 & \citet{cimpoi2013describing}
    \\
    \cmidrule{2-6}
    &   Textures & Final    &  64  & 2,560 & \makecell[l]{\citet{Fritz2004THEKD}\\ \citet{Mallikarjuna2006THEK2} \\ \citet{Kylberg2011c}\\ \citet{lazebnik:inria-00548530}}
    \\
    \midrule
    
    \multirow{2}{*}{Remote sensing}    
    &Mini RSICB &Feedback    &45 &1,800 & \citet{rsicb}
    \\
    \cmidrule{2-6}
     &Mini RESISC &Final   &45 &1,800   & \citet{resisc}
    \\
    \midrule

    \multirow{2}{*}{OCR}    
    &   OmniPrint-MD-mix & Feedback &  706  &  28,240  &  \citet{sun2021omniprint}
    \\
    \cmidrule{2-6}
    &   OmniPrint-MD-5-bis &Final &  706  &  28,240  & \citet{sun2021omniprint}
    \\

    \bottomrule
  \end{tabular}
  \end{adjustbox}
\end{table}

\begin{figure}[ht!]

    \centering
     
    \subfigure[Insects]{
    \includegraphics[width=0.185\linewidth]{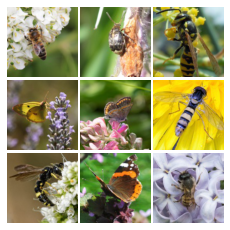}}
    \subfigure[Plant Village]{
    \includegraphics[width=0.185\linewidth]{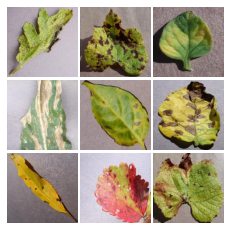}}
    \subfigure[Textures]{
    \includegraphics[width=0.185\linewidth]{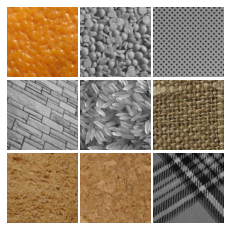}}
    \subfigure[Mini RESISC]{
    \includegraphics[width=0.185\linewidth]{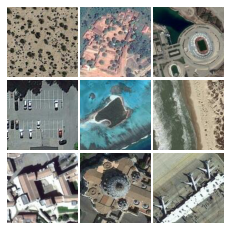}}
    \subfigure[OmniPrint-MD-mix]{
    \includegraphics[width=0.185\linewidth]{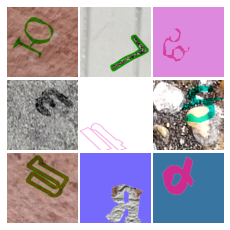}}
    
    \vspace{3ex}
    
    \subfigure[Plankton]{
    \includegraphics[width=0.185\textwidth]{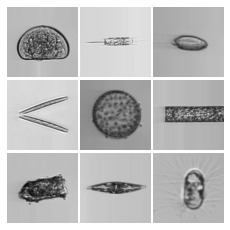}}
    \subfigure[Multiderma]{
    \includegraphics[width=0.185\textwidth]{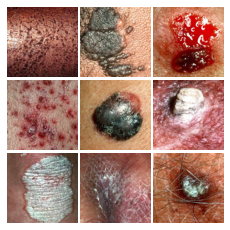}}
    \subfigure[Texture DTD]{
    \includegraphics[width=0.185\textwidth]{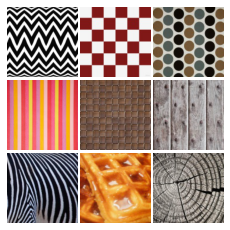}}
    \subfigure[Mini RSICB]{
    \includegraphics[width=0.185\textwidth]{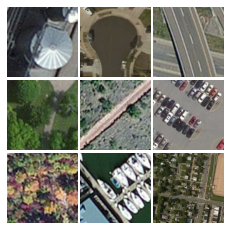}}
    \subfigure[OmniPrint-MD-5-bis]{
    \includegraphics[width=0.185\textwidth]{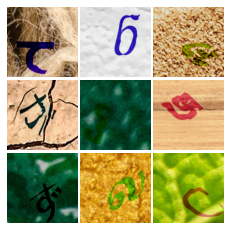}}
    
    \caption{NeurIPS 2021 meta-learning challenge datasets sample images}
    \label{fig:sampleimages}
\end{figure}

\section{Related Work}

While deep neural networks are capable of achieving 
performance superior to humans on various tasks \citep{krizhevsky2012imagenet, mnih2015human, he2015delving}, they are notorious for requiring large amounts of data and processing power, restricting their success to domains where such resources are available.
Humans, on the other hand, are more efficient learners as they can effectively draw on their prior knowledge and learning experience \citep{jankowski2011meta}.
Improving the learning efficiency of deep neural networks is being extensively studied within the area of few-shot learning \citep{wang2020generalizing, bendre2020learning, lu2020learning}.
We discuss the two main paradigms that are used to address this. 

\paragraph{Meta-learning}
Meta-learning \citep{naik1992meta, thrun1998lifelong, schmidhuber1987evolutionary, brazdil2022metalearning} aims to learn, from previous learning experiences, how to learn \citep{vanschoren2018meta, hospedales2020meta, huisman2021survey}. 
Matching networks aim to learn a good embedding such that a nearest-neighbour classifier can be effective \citep{vinyals2016matching}. 
Prototypical networks build on this technique by comparing inputs to class prototypes instead of instances \citep{snell2017prototypical}.
Relation networks replace the distance metric with a neural network \citep{sung2018learning}. 
Model-based approaches, such as MANNs \citep{Santoro16}, Meta-Nets \citep{munkhdalai2017meta}, TURTLE \citep{huisman2021stateless} and SNAIL \citep{mishra2018simple}, embed a given dataset into an activation state and use this state to make predictions for new data points.
Optimization-based approaches use optimization, such as gradient descent, to learn new tasks.
One of the most popular techniques from this approach is MAML \citep{finn2017model}, which aims to learn good initialization hyperparameters from which new tasks can be learned in a few gradient update steps. 
This work has been the inspiration for many follow-up works, such as Meta-SGD \citep{li2017metasgd}, which also learns suitable learning rates, Reptile \citep{nichol2018reptile}, which is a first-order variant on MAML, and LEO \citep{rusu2018meta}, whose goal is
to learn the initialization hyperparameters in a low-dimensional latent space.

\paragraph{Transfer learning}
Transfer learning \citep{weiss2016survey, tan2018survey, pan2009survey} aims to transfer knowledge from a source task or domain (or set thereof), where a large amount of data may be present, to a target domain, where data may be sparse.
One popular transfer learning approach in deep learning is to pre-train a network on a given source domain (e.g., ImageNet \citep{krizhevsky2012imagenet}), followed by fine-tuning parts (such as only the output layer) of the network on the target domain \citep{huang2013cross, oquab2014learning}.   
In this case, the knowledge transfer is parameter-based.
Many other forms of transfer also exist, such as mapping-based, instance-based, and adversarial-based transfer \citep{tan2018survey}.

Recent works illustrate that simple pre-training and fine-tuning can outperform more complicated meta-learning techniques \citep{chen2019closer,tian2020rethinking} which raises the question of whether a good embedding is enough for achieving good few-shot learning performance. 
However, this could also indicate that the few-shot benchmarks such as MiniImageNet \citep{vinyals2016matching, Ravi2017}, TieredImageNet \citep{ren2018meta}, and CUB \citep{wah2011caltech} are not challenging enough because test examples come from the same dataset as the one used for training. 

\paragraph{Related competitions and benchmarks}
This competition is part of an established series of competitions such as the AutoML competition series~\citep{guyon2019analysis}, the AutoDL competition series~\citep{liu_2021}, the AutoCV competition series and the MetaDL competition series~\citep{elbaz2021metadl}.
This competition is an extension to our previous hosted competition in the MetaDL series~\citep{elbaz2021metadl}. It challenges participants with a more challenging set of datasets, that were specifically designed for this challenge. 

The Open Algorithm Selection Competition (OASC) is a competition that is closely related to meta-learning~\citep{lindauer2017open,lindauer2019algorithm}. 
In that competition, for a given dataset, an appropriate algorithm needs to be selected.
While several machine learning datasets are present in the competition, it focuses also on algorithm selection beyond machine learning (e.g., MIP and SAT).

Meta-dataset is another notable benchmark used for few-shot learning. It is a collection of 10 datasets that are commonly used in few-shot learning~\citep{triantafillou2020meta-dataset}. 
Our competition setup with various datasets is partly inspired by this initiative.

}
\end{document}